# Vision-Integrated LLMs for Autonomous Driving Assistance : Human Performance Comparison and Trust Evaluation

Namhee Kim and Woojin Park

*Abstract*— Traditional autonomous driving systems often struggle with reasoning in complex, unexpected scenarios due to limited comprehension of spatial relationships. In response, this study introduces a Large Language Model (LLM)-based Autonomous Driving (AD) assistance system that integrates a vision adapter and an LLM reasoning module to enhance visual understanding and decision-making. The vision adapter, combining YOLOv4 and Vision Transformer (ViT), extracts comprehensive visual features, while GPT-4 enables human-like spatial reasoning and response generation. Experimental evaluations with 45 experienced drivers revealed that the system closely mirrors human performance in describing situations and moderately aligns with human decisions in generating appropriate responses. Semantic similarity metrics, including METEOR and BERT scores, validated the system's reasoning capabilities, while Trust in Automation (TiA) assessments showed increased user trust after interaction. Despite challenges in replicating the diverse cognitive processes of human drivers, this system demonstrates significant potential for augmenting human decision-making in autonomous driving technologies and provides a foundation for further development toward practical deployment.

*Index Terms*—Autonomous driving, computer vision, decision-making, large language models.

## I. INTRODUCTION

### A. Motivations

Autonomous driving (AD) technologies have advanced significantly, demonstrating their transformative potential in transportation systems [8], [11], [30], [50]. The integration of AI has been instrumental in enabling critical functionalities such as perception, decision-making, and motion planning [18], [23], [24], [29], [38]. However, achieving higher levels of autonomy and real-world deployment still faces persistent challenges, particularly in handling complex and unstructured driving scenarios [7], [8], [19], [21].

In recent years, the application of Large Language Models (LLMs) in AD research has gained attention due to their advanced reasoning and common-sense understanding capabilities [1], [9], [15]. Studies have shown that LLMs can support autonomous systems by interpreting multimodal data and generating human-like responses in decision-making processes [49], [53]. Despite this potential, LLM-based systems often struggle with spatial reasoning, which is crucial for navigating dynamic environments [3], [10], [39]. This limitation highlights the need for novel approaches to enhance their ability to understand and act upon spatial relationships in real-world scenarios [58].

To address these challenges, this study proposes a vision-integrated LLM-based AD assistance system and evaluates its effectiveness by comparing its reasoning and decision-making capabilities to those of human drivers. Through this comparison, the study aims to provide insights into how such systems can complement human cognitive processes, enhance trust in autonomous technologies, and advance the broader understanding of AI-assisted decision-making in complex driving scenarios. This research seeks not to replicate human behavior but to explore the potential of vision-integrated LLMs in supporting and augmenting human decision-making in autonomous driving contexts.

### B. Related Works

Autonomous driving systems have long relied on AI-driven advancements to address challenges in perception, decision-making, and control. Deep Neural Networks (DNNs), for instance, have significantly enhanced the capabilities of AD systems in tasks such as object detection, scene understanding, and trajectory prediction [23], [32], [34], [36]. Similarly, Deep Reinforcement Learning (DRL) has proven effective in optimizing motion planning and decision-making processes, enabling vehicles to adapt to dynamic and uncertain environments [6], [25], [27]. These advancements have laid the foundation for the development of Level 1 and Level 2 ADAS, which provide essential safety and convenience features in commercial vehicles [2]. Despite these achievements, higher levels of autonomy present new challenges that require more sophisticated approaches [17], [20], [26].

Recent research has explored the potential of LLMs to enhance the cognitive capabilities of AD systems [33]. LLMs are known for their advanced reasoning abilities, common-sense understanding, and human-like communication skills, making them promising candidates for improving decision-making in autonomous vehicles [1], [12], [16], [53]. Studies have demonstrated their utility in deriving driving decisions through prompt engineering, where textual descriptions of the environment guide the system's responses [43], [45], [51]. Moreover, fine-tuning LLMs on multimodal datasets has further expanded their applicability. For example, DriveGPT4 and RAG-Driver use real-world driving videos to predict throttle and steering angles, while DriveMLM and LMDrive incorporate simulation data to plan trajectories and understand contextual elements [44], [49], [53], [54].

N. Kim is with Seoul National University, Seoul 08826, South Korea. Contact e-mail: nhk506@snu.ac.kr

W. Park is with Seoul National University, Seoul 08826, South Korea. Contact e-mail: woojinpark@snu.ac.kr

However, these LLM-based approaches face critical limitations, particularly in interpreting spatial relationships between objects within visual scenes. The sequential processing nature of LLMs constrains their ability to perform the spatial reasoning required for accurate navigation and decision-making in complex environments [3]. This limitation underscores the importance of integrating vision-based methods with LLMs to overcome these challenges.

Vision-based systems have long been central to autonomous driving, leveraging camera data and other sensors to interpret the surrounding environment [37], [46]. Traditional approaches augmented by DNNs have significantly improved tasks like object detection, segmentation, and scene analysis [23], [29], [35], [42], [47], [48]. However, these methods often operate independently of decision-making processes, leading to inefficiencies in handling complex driving scenarios. To address this, researchers have proposed combining vision adapters with LLMs to enhance spatial reasoning and integrate visual and textual data more effectively. Vision adapters, such as YOLOv4 and Vision Transformer (ViT), extract comprehensive visual features that improve the system's ability to interpret and respond to dynamic driving environments [4], [5].

By integrating these advanced vision-based methods with the reasoning capabilities of LLMs, researchers aim to bridge the critical gaps in existing AD systems. This approach not only enhances spatial comprehension but also aligns autonomous systems more closely with human cognitive processes, paving the way for safer and more adaptable autonomous driving technologies.

*C. Contribution*

This study examines the potential of a vision-integrated LLM-based framework for AD assistance systems to address challenges in spatial reasoning and decision-making in complex driving environments. By integrating a vision adapter, which combines YOLOv4 and ViT, with an LLM reasoning module, the proposed system enhances its ability to interpret spatial relationships and generate contextually appropriate responses. The focus of this research is on evaluating its performance in comparison to human drivers to explore its alignment with human cognitive processes.

To this end, the study compares the outputs of the vision-integrated LLM system with human reasoning across scenarios involving unexpected driving situations. Using experimental designs that involve image-video pairs, the research investigates the system's ability to describe situations accurately and make decisions that are consistent with human responses. By analyzing similarities and differences, this study provides insights into how such systems can support and augment human decision-making rather than replicate it.

This research emphasizes the broader implications of integrating vision and language models in autonomous driving, contributing to a deeper understanding of how these systems can enhance trust and reliability in real-world applications. Through its evaluation-driven approach, the study lays the groundwork for designing AD assistance systems that better align with human cognitive expectations, fostering safer and more adaptive autonomous technologies.

*D. Paper Organization*

The remainder of this paper is organized as follows.
Section II describes the methodologies, focusing on the design of the model architecture, the datasets used, and the training process. Section III outlines the experimental setup and data collection methods, including participant tasks and the procedures for gathering and preparing data. Section IV presents the results and analysis, covering system performance, human-AI response similarity, and the impact of the system on user trust levels. Section V discusses the implications of the findings and highlights areas for future work. Finally, the paper is concluded in Section VI.

## II. METHODOLOGIES

This section describes the model framework that is used in this study. Specifically, Section II-A presents the system's architecture, integrating vision and language components, while Section II-B outlines the training process, including preprocessing and evaluation strategies.

*A. Model Architecture and Datasets*

The proposed LLM-based autonomous driving assistance system consists of two core components: a vision adapter and a LLM. The vision adapter processes visual inputs and extracts relevant features, which are then used by the LLM to generate situation descriptions and determine appropriate responses.

The vision adapter integrates YOLOv4 for object detection and a pre-trained ViT for spatial relationship analysis [22], [31], [55]. YOLOv4 employs a grid-based approach to detect objects, while ViT divides the input image into patches and identifies relationships between them. The extracted features are aligned with the LLM's embedding space using a linear projection layer, facilitating seamless interaction between the visual and language components [56].

The LLM, based on GPT-4, is used as the reasoning module to convert visual features into text-based outputs. During deployment, the temperature parameter of GPT-4 was set to 0.7, balancing creativity and precision in the generated responses. A higher temperature (e.g., >0.9) was avoided as it could lead to overly varied and inconsistent outputs, while a lower temperature (<0.5) risked generating overly deterministic responses, limiting adaptability to diverse driving scenarios. The selected temperature ensured the system provided coherent yet contextually flexible outputs, aligning with the requirements for autonomous driving scenarios.

For training, the Berkeley DeepDrive (BDD100k) dataset was utilized, offering a diverse range of real-world driving scenarios, including variations in geography, weather, and traffic conditions [52]. This dataset contains annotated data for 10 object classes, such as cars, pedestrians, traffic lights, and bicycles, making it suitable for both object detection and contextual reasoning tasks. The dataset's comprehensive coverage of urban and highway driving scenarios allowed the model to generalize effectively to various environments.

## B. Model Training

The model training process adhered to a supervised learning framework, where the system was trained to map input data to corresponding outputs. To ensure robust evaluation and prevent data leakage, the dataset was divided into training (80%) and validation (20%) subsets. This commonly used partitioning ratio balances model optimization during training with performance assessment on unseen data [14], [57].

Before training, the dataset underwent preprocessing to standardize image pixel values to a [0, 1] range. To improve the model's robustness and generalizability, data augmentation techniques such as flipping, rotation, cropping, and brightness adjustment were applied. These steps ensured that the model could adapt to diverse driving scenarios, including variations in lighting, weather, and traffic density.

The training process employed the Adam optimizer, with the initial learning rate set to 0.001, enabling efficient weight updates for object detection tasks [40]. YOLOv4, designed for grid-based object detection, was trained using this setup, while the ViT, pre-trained on ImageNet, was fine-tuned with the BDD100k dataset to adapt its spatial reasoning capabilities to real-world driving environments [13]. Batch normalization and dropout techniques were applied throughout training to mitigate overfitting, ensuring the model's generalizability across diverse inputs.

The integration of the Vision Adapter and GPT-4 was achieved by aligning visual features with the language model's embedding space through a linear projection layer. This alignment allowed GPT-4 to effectively reason about spatial relationships and generate contextually appropriate outputs. The temperature parameter of GPT-4 was set to 0.7 during deployment, striking a balance between creativity and precision in the generated responses. Lower temperatures (<0.5) were avoided to ensure adaptability in diverse scenarios, while higher temperatures (>0.9) were avoided to maintain response consistency.

Model performance was continuously monitored during training using standard metrics, including precision, recall, and F1-score. Validation was conducted after every epoch to assess the model's ability to generalize to unseen data. To prevent overfitting and ensure optimal convergence, early stopping was implemented, halting training if validation loss did not improve for five consecutive epochs.

By combining rigorous preprocessing, robust training methodologies, and a high-performance computing environment, the system was optimized for accurate and reliable operation in complex driving scenarios.

## III. EXPERIMENT AND DATA COLLECTION

This section describes the experimental design and data processing for evaluating the proposed system's performance and its impact on user trust. It outlines the experimental procedure, including participant tasks and assessments, as well as the methods for data collection and analysis.

## A. Experimental Procedure and Participants

The experimental procedure was designed to evaluate the performance of the proposed system and its influence on user trust. Participants were invited to interact with the system by observing visual stimuli, including images and videos depicting unexpected and challenging driving scenarios, and performing cognitive reasoning tasks based on the observed content. The experiment was structured into three sequential stages to ensure a comprehensive assessment of user trust and system performance.

First, participants underwent an Initial Trust Assessment to establish baseline trust levels in the system. For this purpose, they completed the Trust in Automation (TiA) scale [28], a validated instrument designed to evaluate multiple dimensions of trust in automated systems. The TiA scale measures trust across six components: reliability/competence, understanding/predictability, familiarity, intention of developers, propensity to trust, and overall trust in automation. By addressing these dimensions, the TiA scale provided a nuanced understanding of participants' initial trust levels, offering a baseline against which changes in trust could be measured throughout the experiment.

Next, during the Task Execution phase, participants were presented with three distinct driving scenarios, each representing a unique and high-stakes situation designed to reflect real-world challenges. The first scenario, An Abrupt Lane Change Necessitated by an Accident Ahead, involved a sudden maneuver required to avoid a blocked lane caused by a traffic accident. The second scenario, A Vehicle Rollover Accident due to a Sudden Obstacle, described a situation where a vehicle lost control and overturned after encountering an unexpected obstacle. The third scenario, A Multi-Vehicle Collision Caused by Severe Weather Conditions, depicted a chain-reaction crash initiated by low visibility and hazardous weather. For each scenario, participants were instructed to provide a detailed situation description, outlining their observations of the key elements and context of the event. In addition, they were required to articulate at least two appropriate responses for each case, proposing effective strategies to address the challenges presented. These responses needed to draw on their personal driving experience and consider both immediate actions and long-term preventative measures.

Finally, in the Final Trust Assessment, participants were shown the AI-generated responses for the same driving scenarios they had previously analyzed. After reviewing the system's outputs, they completed a post-trust evaluation using the TiA scale to reassess their trust levels. This step facilitated a direct comparison between participants' initial trust levels and their perceptions after interacting with the system's outputs.

A total of 45 participants were recruited, comprising both men and women with a minimum of five years of driving experience. The demographic distribution of participants is summarized in Table I. Ethical approval for the study was granted by the Institutional Review Board of Seoul National University (Protocol 2407/004-009), and informed consent was obtained from all participants before the experiment.

TABLE I
PARTICIPANT DEMOGRAPHIC INFORMATION

| Age Group | Gender Ratio | Years of Driving Experience |
|---|---|---|
| 20s (n=13) | 10 females, 3 males | M=5.85, SD=1.03 |
| 30s (n=13) | 2 females, 11 males | M=8.46, SD=2.56 |
| 40s (n=8) | 4 females, 4 males | M=13.25, SD=4.63 |
| 50s (n=7) | 3 females, 4 males | M=16.29, SD=7.44 |
| 60s (n=4) | 4 males | M=33.5, SD=2.60 |

*B. Experimental Data Collection*

Experimental data were collected to evaluate the system's reasoning capabilities and its effect on user trust. Participants generated textual situation descriptions and appropriate responses for three driving scenarios. These responses were compared with the system's outputs to assess semantic similarity [41].

To measure trust levels, the TiA scale was administered both before and after the experiment. The TiA scale consists of 19 items covering five dimensions: reliability/competence, familiarity, trust, understanding, and intention of developers. Each item was rated on a 5-point Likert scale ranging from 1 (strongly disagree) to 5 (strongly agree). This provided a quantitative measure of changes in trust resulting from interaction with the system.

IV. EXPERIMENT RESULTS AND ANALYSIS

This section presents the results and analysis of the experiment, focusing on system performance, human-AI response similarity, and changes in user trust. Key findings include the evaluation of the vision adapter's performance, the semantic alignment of AI-generated outputs with human reasoning, and statistical analysis of trust level differences.

*A. Experimental Data Analysis*

The analysis of experimental data was conducted using the following approaches:

*1) Vision Adapter Performance*: The vision adapter's object detection and classification capabilities were evaluated using precision, recall, and F1-score, which are standard performance metrics in computer vision [14]. These metrics provided insights into the system's ability to accurately identify and describe objects within driving scenarios. Data analysis for this evaluation was performed using Python libraries to compute the performance metrics.

*2) Human-AI Similarity*: Two independent experts assessed the semantic similarity between human-written and AI-generated responses using a 5-point Likert scale. This evaluation determined how closely the system's outputs aligned with human reasoning, with scores ranging from 1 (completely dissimilar) to 5 (highly similar).

*3) Trust Level Analysis*: Changes in trust levels were analyzed using paired t-tests, comparing pre- and post-experiment TiA scores. Statistical significance was determined at a 95% confidence level ($p < 0.05$). Effect sizes (Cohen's d) were calculated to assess the magnitude of trust changes, providing additional insights into the system's influence on user confidence [16]. Data analysis was performed using Python libraries and SPSS.

*B. Evaluation of Vision Adapter Performance*

The proposed LLM-based autonomous driving assistance system was tested on three unexpected driving scenarios: abrupt lane changes due to accidents ahead (case 1), vehicle rollovers caused by sudden obstacles (case 2), and multi-vehicle collisions under severe weather conditions (case 3). Each scenario was selected to evaluate the system's ability to generate accurate situation descriptions and appropriate responses.

The vision adapter, combining YOLOv4 and ViT, demonstrated high accuracy in object detection and spatial relationship analysis. For object detection, the adapter achieved a precision of 89.5%, recall of 91.2%, and an F1-score of 90.3% (Table II). These metrics indicate the system's robust performance in identifying objects such as "white pickup truck" and "cargo truck" and providing detailed descriptions including directions "left," "right," and "front".

TABLE II
PERFORMANCE OF VISION ADAPTER OBJECT DETECTION

| Precision (%) | Recall (%) | F1-score (%) |
|---|---|---|
| 89.5 | 91.2 | 90.3 |

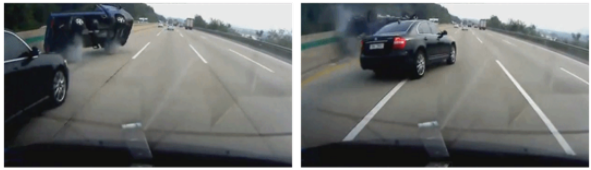

An Abrupt Lane Change Necessitated by An Accident Ahead (Case 1)

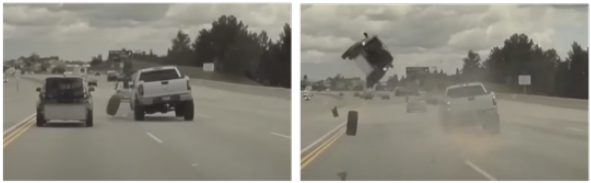

A Vehicle Rollover Accident due to a Sudden Obstacle (Case 2)

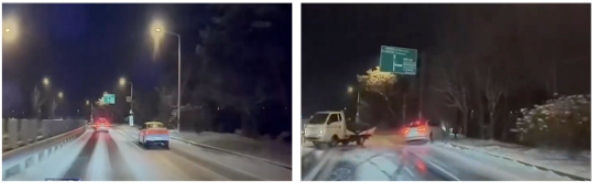

A Multi-Vehicle Collision caused by Severe Weather Conditions (Case 3)

Fig. 1. Model Response Results by Cases

Fig. 1. illustrates the situation descriptions and appropriate responses derived by the AI system for the three scenarios. While the situation descriptions accurately detailed objects and their locations, the appropriate responses included common-sense guidelines, such as slowing down and maintaining a safe distance when encountering an accident ahead. These outputs align closely with expected human reasoning and demonstrate the system's capacity for practical decision-making.

## C. Evaluation of Human-AI Response Similarity

To evaluate the similarity between human and AI responses, two independent experts analyzed the textual outputs for 45 participants across three scenarios. Each participant provided both a situation description and an appropriate response, yielding a total of 540 evaluation scores (45 participants × 3 images × 2 response categories × 2 experts).

The analysis revealed an average similarity score of 4.20 (SD = 0.80, SE = 0.05) for situation descriptions, indicating strong alignment with human-generated outputs. For appropriate responses, the average similarity score was 3.38 (SD = 0.95, SE = 0.06), suggesting moderately lower agreement (Table III). Fig. 2. visualizes these results, emphasizing the AI system's strength in visual recognition while also highlighting its reliance on standardized responses. This reliance limits the variability and contextual adaptability observed in human responses.

TABLE III
SIMILARITY SCORES BY RESPONSE CATEGORY

|  | M (SD) | SE |
| --- | --- | --- |
| Situation Description | 4.20 (0.80) | 0.05 |
| Appropriate Response | 3.38 (0.95) | 0.06 |

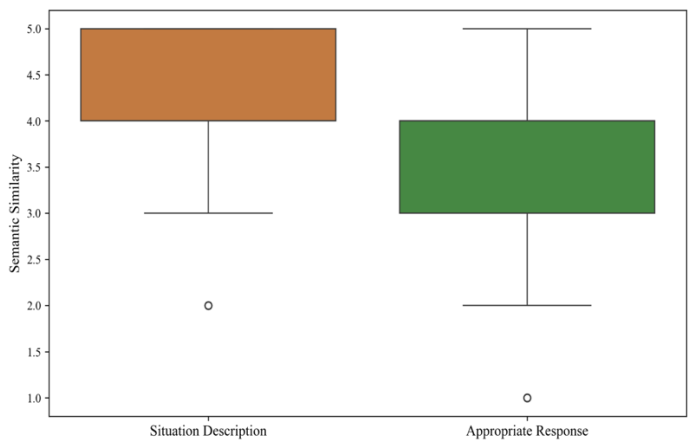

Fig. 2. Boxplot of Response Category

To further analyze the quality of the AI-generated textual outputs, we incorporated METEOR and BERT score metrics to provide a quantitative evaluation of similarity (Table IV). For situation descriptions, the AI achieved a METEOR score of 0.76 (SD = 0.09), reflecting strong semantic alignment through recognition of synonyms and paraphrases. The corresponding BERT score was 0.73 (SD = 0.11), demonstrating the system's ability to understand contextual relationships and nuanced language.

For appropriate responses, the METEOR score was 0.75 (SD = 0.13), comparable to that of situation descriptions but with slightly greater variability. The BERT score for this category was 0.75 (SD = 0.09), indicating a consistent performance in maintaining contextual relevance, although the system's standardized outputs still lack the creative variability often seen in human responses.

TABLE IV
EVALUATION METRICS OF SEMANTIC SIMILARITY FOR AI-GENERATED TEXT

|  | Situation Description | Appropriate Response |
| --- | --- | --- |
|  | M (SD) | M (SD) |
| METEOR score | 0.76 (0.09) | 0.75 (0.13) |
| BERT score | 0.73 (0.11) | 0.75 (0.09) |

Table V provides additional insights by examining the User Narrative Data with the highest similarity scores for specific scenarios. In cases where spatial relationships and object details were critical, the system demonstrated robust performance, aligning closely with human reasoning in situation descriptions.

However, in the appropriate response tasks, the scores were lower due to the AI system's standardized outputs, which contrast with the creative and diverse solutions based on personal experiences often seen in human-generated narratives. These findings underline the system's strengths in structured tasks while pointing to areas for improvement in handling diverse and context-rich scenarios.

TABLE V
USER NARRATIVE DATA ACHIEVING THE HIGHEST SCORES BY CASE/CATEGORY

| Case | Categories | User Narrative Data | Similarity Scores (Expert A) | Similarity Scores (Expert B) |
|---|---|---|---|---|
| Case 1 | Situation Description | The current situation involves an overturned vehicle in the first lane of a single-lane road. Vehicles traveling in the first lane are quickly merging into the second lane to avoid the accident. Although lane changes are prohibited due to the solid line between the two lanes, the urgency of the situation forces the vehicles in the first lane to merge into the second lane abruptly, creating the risk of a secondary collision with you, the driver. | 5 | 5 |
| | Appropriate Response | In such a situation, the best response is to quickly recognize the accident from a distance, turn on the hazard lights, and alert surrounding vehicles to the current situation. Additionally, you should anticipate that vehicles from the first lane might merge into the second lane. Reduce your speed to allow the merging vehicles to enter safely, thereby preventing secondary accidents. | 5 | 5 |
| Case 2 | Situation Description | A truck traveling in the second lane suddenly loses a wheel, which rolls into the first lane and hits a passenger car, causing a severe rollover accident. | 5 | 5 |
| | Appropriate Response | This situation poses a significant threat to the lives of the driver and passengers in the accident vehicle. Additionally, vehicles following the accident car may encounter rolling debris or the detached wheel, leading to secondary accidents. To mitigate further incidents, it is crucial to immediately turn on the hazard lights to alert following vehicles of the accident. You should also brake promptly and, if possible, change lanes to the right to avoid additional collisions. On highways, maintaining sufficient distance between vehicles, adhering to speed limits, avoiding reckless overtaking, and ensuring a pre-trip vehicle inspection are essential practices for safety. | 5 | 5 |
| Case 3 | Situation Description | The current situation involves heavy snowfall and low temperatures, where snow that melted during the day has turned into ice. A truck skids on the icy road, blocking both the first and second lanes. A car traveling slowly in the second lane attempts an emergency stop to avoid a collision but veers off into the median strip. | 5 | 5 |
| | Appropriate Response | In such conditions, the best course of action is to turn on the hazard lights immediately to alert surrounding vehicles to the slippery road conditions caused by the icy snow. At the same time, ensure that you maintain a safe following distance and drive at reduced speeds. Using tire spray chains or attaching snow chains to the tires before driving are also effective measures for enhancing safety in these conditions. | 5 | 5 |

*D. Analysis of Trust Level Differences*

To measure changes in user trust, pre- and post-experiment scores were collected using the TiA scale, consisting of 19 items rated on a 5-point Likert scale. A paired samples t-test revealed a statistically significant increase in trust, with mean scores rising from 50.70 (SD = 12.18) to 59.97 (SD = 15.16) (t(44) = 5.030, p < .001) (Table VI). This corresponds to an average trust increase of 9.27 percentage points (95% CI [5.55, 12.98]), demonstrating the AI system's positive influence on user confidence (Table VII). Effect size analysis further supported the significance of this improvement, with a Cohen's d value of 0.750 (95% CI [0.415, 1.078]) and Hedges' g of 0.737 (95% CI [0.408, 1.059]) (Table VIII). Additionally, a moderate positive correlation was observed between pre- and post-experiment trust levels (r = 0.610, p < .001), indicating that participants with higher initial trust were more likely to maintain or improve their confidence. This trend is visualized in Fig. 3. No significant correlations were found between trust levels and demographic variables such as age or gender, suggesting consistent trust improvements across diverse participant groups. These findings highlight the potential of user-friendly AI systems to foster trust, particularly in safety-critical applications like autonomous driving.

TABLE VI
MEAN TRUST LEVELS WITH PAIRED t-TEST RESULTS

|  | M (SD) | SE |
|---|---|---|
| Final Trust Level | 59.97 (15.16) | 2.26 |
| Initial Trust Level | 50.70 (12.18) | 1.81 |

TABLE VII
STATISTICAL SUMMARY OF TRUST LEVEL DIFFERENCES

|  | M (SD) | SE | 95% Confidence Interval of the Difference | | t | df | Sig. (2-tailed) |
|---|---|---|---|---|---|---|---|
|  |  |  | Lower | Upper |  |  |  |
| Final Trust Level − Initial Trust Level | 9.27 (12.36) | 1.84 | 5.54 | 12.98 | 5.03 | 44 | .000 |

TABLE VIII
EFFECT SIZE ANALYSIS OF TRUST LEVEL DIFFERENCES

|  |  | Standardizer[a] | Point estimate | 95% CI | |
|---|---|---|---|---|---|
|  |  |  |  | Lower | Upper |
| Final Trust Level − Initial Trust Level | Cohen's d | 12.36 | .750 | .415 | 1.078 |
|  | Hedges' g | 12.58 | .737 | .408 | 1.059 |

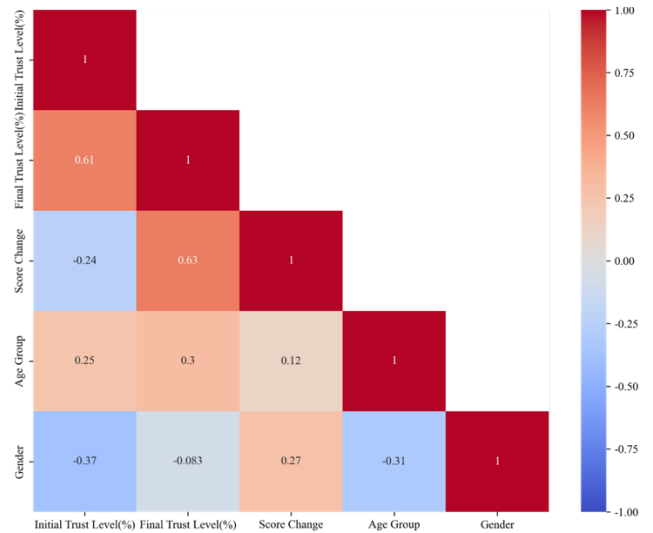

Fig. 3. Correlation Matrix for Trust Evaluation

V. DISCUSSION

This study developed and evaluated a vision-integrated LLM-based AD assistance system, demonstrating its potential to enhance reasoning processes in complex driving scenarios. The findings revealed that the system achieved high similarity to human performance in situation description tasks, showcasing its ability to effectively recognize and interpret objects and their spatial relationships in visual scenes. However, the system demonstrated moderately lower similarity in generating appropriate responses, highlighting the inherent complexity and variability of human decision-making, which the current system has yet to fully replicate.

The gap in performance between situation descriptions and appropriate responses underscores the challenges of developing AI systems capable of handling diverse and dynamic contexts. While the integration of LLMs has significantly improved reasoning capabilities, the system's reliance on standardized outputs limits its adaptability to unpredictable and context-specific situations. Future research should prioritize enhancing the system's contextual reasoning by incorporating reinforcement learning techniques and expanding training datasets to include a wider variety of scenarios. These advancements will enable the system to handle more diverse situations and provide responses that better align with human decision-making processes.

Another key finding of this study was the system's positive impact on user trust. Interaction with the system resulted in a statistically significant increase in trust levels, demonstrating its ability to enhance user confidence in autonomous technologies. A moderate positive correlation between initial and final trust scores indicates that participants with higher initial trust levels experienced greater confidence gains. Importantly, no significant correlations were observed between trust and demographic factors such as age or gender, suggesting the system's universal applicability across diverse participant groups.

Despite these promising results, the study has several limitations that warrant attention in future research. First, the experiments were conducted in a controlled environment with a limited number of predefined scenarios, restricting the generalizability of the findings. Testing the system under real-world driving conditions is critical to evaluate its robustness and reliability in practical applications. Second, the system's dependence on existing training datasets may limit its performance in rare or extreme scenarios not covered during training, posing potential risks in critical situations. Third, while this study utilized GPT-4 as the LLM reasoning module, the rapid advancements in LLM technologies may render this version obsolete as newer versions are released. As such, future systems may need regular updates to leverage the improved capabilities of next-generation LLMs and maintain competitive performance. Moreover, ensuring real-time decision-making under varying and complex traffic conditions remains a significant hurdle. Addressing latency and computational efficiency in real-world deployments will be crucial to enhancing the practical usability of such systems. Furthermore, the ethical challenges associated with AI decision-making in ambiguous or conflicting situations remain unresolved, requiring further exploration to ensure safe and effective integration into human-centric applications.

In conclusion, the findings of this study offer valuable insights into the development of LLM-based AD systems, particularly in enhancing situational awareness and fostering user trust. However, the limitations identified underscore the importance of careful and deliberate efforts in advancing these technologies. By addressing the critical challenges of real-time reliability and achieving absolute safety (zero accident tolerance), which are essential for commercial viability, future research can focus on practical solutions that harmonize technological innovation with ethical considerations. These efforts will advance the development of autonomous driving systems that are not only reliable and adaptable but also aligned with human-centric values.

## VI. CONCLUSION

This study successfully developed and evaluated an LLM-based AD assistance system that integrates GPT-4 with a vision adapter, demonstrating its potential to contribute to the development of autonomous driving assistance systems. The system excelled in structured tasks, particularly in situation description, effectively recognizing and interpreting spatial relationships in visual scenes. However, its lower performance in generating appropriate responses highlights the need for further refinement to handle diverse and context-rich scenarios.

The findings underline the system's strengths in structured reasoning but also reveal its reliance on standardized outputs, which limits its adaptability to dynamic and unpredictable contexts. Future research should prioritize improving the system's contextual reasoning by expanding training datasets and incorporating advanced techniques to better address the variability of human decision-making.

These results provide a foundation for advancing LLM-based AD assistance systems. By focusing on real-world validation, leveraging advancements in LLM technologies, and addressing ethical considerations, future iterations can help develop more adaptable, reliable, and human-centered autonomous driving assistance technologies.

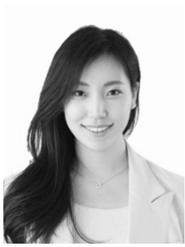

**Namhee Kim** received the B.S. degree in information security from Korea University, Seoul, South Korea, in 2022. She is currently pursuing the M.S. degree in industrial engineering at Seoul National University, Seoul, South Korea. Her research interests include AI, user experience, and digital human modeling.

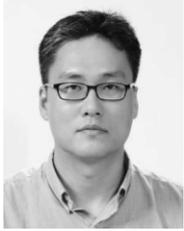

**Woojin Park** received the B.S. and M.S. degrees in industrial engineering from the Pohang University of Science and Technology, Pohang, South Korea, in 1995 and 1997, respectively, and the Ph.D. degree in industrial and operations engineering from the University of Michigan, Ann Arbor, in 2003. He is currently a Professor of industrial engineering with Seoul National University, Seoul, South Korea. His research interests include vehicle ergonomics, creativity techniques, ergonomics design for obese individuals, and digital human modeling.